# A Survey on Sentiment and Emotion Analysis for Computational Literary Studies

Evgeny Kim and Roman Klinger

Institut für Maschinelle Sprachverarbeitung
University of Stuttgart

evgeny.kim@ims.uni-stuttgart.de
roman.klinger@ims.uni-stuttgart.de

Abstract

Emotions are a crucial part of compelling narratives: literature tells us about people with goals, desires, passions, and intentions. Emotion analysis is part of the broader and larger field of sentiment analysis, and receives increasing attention in literary studies. In the past, the affective dimension of literature was mainly studied in the context of literary hermeneutics. However, with the emergence of the research field known as Digital Humanities (DH), some studies of emotions in a literary context have taken a computational turn. Given the fact that DH is still being formed as a field, this direction of research can be rendered relatively new. In this survey, we offer an overview of the existing body of research on emotion analysis as applied to literature. The research under review deals with a variety of topics including tracking dramatic changes of a plot development, network analysis of a literary text, and understanding the emotionality of texts, among other topics.

## 1 Introduction and Motivation

This article deals with *emotion* and *sentiment* analysis in *computational literary studies*. Following Liu[1], we define sentiment as a *positive* or *negative* feeling underlying the opinion. Sometimes, sentiment analysis is interpreted synonymously to opinion mining, however strictly speaking, opinion mining is an application that makes use of sentiment analysis and contextualizes polarity ratings in topics, aspects and targets. Though sentiment analysis is primarily text-oriented, there are multimodal approaches as well.[2]

Another interpretation of the term *sentiment analysis* is as broader description of a research field, which considers affective computing applied to textual analysis. In this sense, it also includes the distinction into subjective or objective statements[3], and, more recently, the field of emotion analysis. Defining the concept of *emotion* is a challenging task. As Scherer puts it, defining emotion is a notorious problem.[4] Indeed, different methodological and conceptual approaches to dealing with emotions lead to different definitions. However, the majority of emotion theorists agree that emotions involve a set of expressive, behavioral, physiological, and phenomenological features.[5] In this view, an emotion can be defined as an integrated feeling state involving physiological changes, motor-preparedness, cognitions about action, and inner experiences that emerges from an appraisal of the self or situation.[6]

[1] Liu 2015, p.2.
[2] Soleymani et al. 2017, passim.
[3] Wiebe et al. 2004, passim.

Similar to sentiment, emotions can be analyzed computationally. However, the goal of emotion analysis is to recognize the emotion, rather than sentiment, which makes it a more difficult task as differences between some emotion classes are more subtle than those between positive and negative.

Although sentiment and emotion analysis are different tasks, our review of the literature shows that the use of either term is not always consistent. There are cases where researchers analyze only positive and negative aspects of a text but refer to their analysis as emotion analysis. Likewise, there are cases where researchers look into a set of subjective feelings including emotions but call it sentiment analysis. Hence, to avoid confusion, in this survey, we use the terms *emotion analysis* and *sentiment analysis* interchangeably. In most cases, we follow the terminology used by the authors of the papers we discuss (i.e., if they call emotions sentiments, we do the same). However, our focus of this survey is on emotion analysis, and we do not include the majority of work that focuses on binary polarities.

Finally, we talk about sentiment and emotion analysis in the context of computational literary studies. Da defines computational literary studies as the statistical representation of patterns discovered in text mining fitted to currently existing knowledge about literature, literary history, and textual production.[7] Computational literary studies are closely related to the concepts of *distant reading*[8] and *digital literary studies*,[9] each of which refers to the practice of running a textual analysis on a computer to yield quantitative results. In this survey, we use all of these terms interchangeably and when we refer to digital humanities as a field, we refer to those groups of researchers whose primary objects of study are texts.

## 1.1 Scope of this Survey

This survey provides an overview of work which aims at understanding or analyzing emotions in literature. We include studies that answer a concrete research question from the field of literary studies with computational methods. We do only consider publications in English that have been quality-assessed by peer review (except for few exceptions). We exclude efforts of corpus creation and annotation, if those corpora have not been used for a further research study to limit the scope of this survey (though such work is clearly relevant

4 Scherer 2005, p. 1.
5 Scarantino 2016, p. 36.
6 Mayer et al. 2008, p. 2.
7 Da 2019, p. 602.
8 Moretti 2005, passim.
9 Hoover et al. 2014, passim.

and important) and software development efforts if the associated papers do not aim at contributing to answering a research question. Similarly, we do mostly exclude reports on ongoing research efforts, if they do not contribute a novel understanding of a research question. Our literature research started in the field of computational linguistics with the ACL Anthology[10] and has been complemented by other research that cites such papers or is cited by them. We exclude papers from local digital humanities conferences.

The goal of this survey is to provide an overview of recent methods of emotion and sentiment analysis as applied to a text. The survey is directed at researchers looking for an introduction to the existing research in the field of sentiment and emotion analysis of a (primarily, literary) text. We do not not cover applications of emotion analysis in the areas of digital humanities that are not focused on text. Neither do we provide an in-depth overview of all possible applications of emotion analysis in the computational context outside of the DH line of research.

## 1.2 Emotion Analysis and Digital Humanities

Methods that apply emotion analysis can in general be categorized into (1) dictionary-based methods, (2) feature-based machine-learning-based, and (3) representation-learning/deep learning-based. Methods that apply statistical learning (2,3) to induce a model that takes text as input and outputs predictions rely in the majority of cases (in this field) on supervised approaches – a learning algorithm is presented with annotated data and needs to output a model that can, as good as possible on unseen data, do such predictions. These approaches have advantages: The learner can exploit (long-distant) dependencies between textual units, learn associations between semanic meaning and concepts to learn, and make use of semantic similarities between words; even those that have not been seen in training data. This comes at a cost – the need for annotated data. The situation between the fields of computational linguistics and digital humanities differs substantially in this regard.

The focus in computational linguistics is to develop methods to solve a particular task – analyze syntax, respresent semantics, or develop well-performing classification methods, for instance for emotion classification. Therefore, there exists a substantial body of research on natural language processing which is essentially agnostic to the corpus. In fact, a method is typically evaluated on a set of different resources to prove its generalizability, and even if a novel corpus is presented for future studies, this is compared to existing resources. This comes with an advantage: Resources are often built by domain experts, which are then used for further analysis; the diversity might be limited, but is often sufficient for model development.

In digital humanities, this situation differs substantially. The goal is often not the development of a computational model that is able to make predictions for the entirety of a field (which is of course also not achieved in computational linguistics, but that is sometimes claimed to be a goal). Instead, the object of research (a particular text, a genre, an author, ...) is of higher importance. This comes with a challenge: Annotators often need to be experts in the particular domain, for a particular object of research.

---

[10] https://www.aclweb.org/anthology/

That might be the reason, as we will see, that, in contrast to research in computational linguistics, using lexicons of words associated with the concepts of interest, receives some attention as a methodological approach to emotion analysis. This comes at the cost of accuracy, as such methods are (mostly) not able to interpret the context appropriately (with some exceptions which embed dictionaries with rules[11]), however, it contributes the advantage of being transparent not only with the predictions and the results, but also with the analysis algorithm.

## 1.3 Emotions and Arts

Much of our daily experiences influence and are influenced by the emotions we experience.[12] This experience is not limited to real events. People can feel emotions because they are reading a novel or watching a play or a movie.[13] There is a growing body of literature that pinpoints the importance of emotions for literary comprehension,[14] as well as research that recognizes the deliberate choices people make with regard to their emotional states when seeking narrative enjoyment such as a book or a film[15].

The link between emotions and arts in general is a matter of debate that dates back to the Ancient period, particularly to Plato, who viewed passions and desires as the lowest kind of knowledge and treated poets as undesirable members in his ideal society.[16] In contrast, Aristotle's view on emotive components of poetry expressed in his *Poetics*[17] differed from Plato's in that emotions do have great importance, particularly in the moral life of a person.[18] In the late nineteenth century the emotion theory of arts stepped into the spotlight of philosophers. One of the first accounts on the topic is given by Leo Tolstoy in 1898 in his essay *What is Art?*.[19] Tolstoy argues that art can express emotions experienced in fictitious context and the degree to which the audience is convinced of them defines the success of the artistic work.[20]

New methods of quantitative research emerged in humanities scholarship bringing forth the so-called *digital revolution*[21] and the transformation of the field into what we know as digital humanities.[22] The adoption of computational methods of text analysis and data mining from the fields of then fast-growing areas of computational linguistics and artificial intelligence provided humanities scholars with new tools of text analytics and data-driven approaches to theory formulation.[23]

To the best of our knowledge, the first work[24] on a computer-assisted modeling of emotions in literature appeared in 1982. Challenged by the question of why some texts are more interesting than others, Anderson and McMaster concluded that the emotional tone of a story can be responsible for the reader's interest. The results of their study suggest that a large-scale analysis of the emotional tone of a collection of texts is possible with the help of a computer program. There are two implications of this finding. First, they suggested that by identifying emotional tones of text passages one can model affective patterns of a given text or a collection of texts, which in turn can be used to challenge or test existing literary theories. Second, their approach to affect modeling demonstrates that the stylistic properties of texts can be defined on the basis of their emotional interest and not only their linguistic characteristics. With regard to these implications, this work is an important early piece as it laid out a roadmap for some of the basic applications of sentiment and emotion analysis of texts, namely sentiment and emotion pattern recognition from text and computational text characterization based on sentiment and emotion.

---

[11] e.g., Shaikh 2009, passim.
[12] Schwarz 2000, p. 1.
[13] Johnson-Laird / Oatley 2016, passim; Djikic et al. 2009, passim.
[14] Robinson 2005; Hogan 2010; Hogan 2011; Bal / Veltkamp 2013; Djikic et al. 2013; Johnson 2012; Samur et al. 2018.
[15] Zillmann et al. 1980; Ross 1999; Bryant / Zillmann 1984; Oliver 2008; Mar et al. 2011.
[16] Plato 1969, passim.
[17] Aristotle 1996, passim.
[18] de Sousa / Scarantino 2018, passim.
[19] Tolstoy 1962, passim.
[20] Anderson / McMaster 1986, passim; Hogan 2010, passim; Piper / Jean So 2015, passim.
[21] Lanham 1989, passim.
[22] Berry 2012, passim; Schreibman et al. 2015, passim.
[23] Vanhoutte 2013, passim; Jockers / Underwood 2016, passim.
[24] Anderson / McMaster 1982, passim.

With the development of research methods used by digital humanities researchers, the number of approaches and goals of emotion and sentiment analysis in literature has grown.

## 2 Affect and Emotion

The history of emotion research has a long and rich tradition that followed Darwin's 1872 publication of *The Expression of the Emotions in Man and Animals*[25]. The subject of emotion theories is vast and diverse. We refer the reader to Maria Gendron's paper[26] for a brief history of ideas about emotion in psychology. Here, we will focus on three views on emotion that are popular in computational analysis of emotions (though they are, from a psychological perspective, motivated from different perspectives and represent different elements of affect and emotion): Ekman's *theory of basic emotions*, Plutchik's *wheel of emotion*, and Russel's *circumplex model*.

## 2.1 Ekman's Theory of Basic Emotions

The idea of basic emotion theories is that there are emotions that are more "fundamental" than others. Mixtures of emotions which receive a particular other name are not necessarily defined as being basic. These attempts to find an emotion definition date back to Silvan Tomkins[27] in the early 1960s, who characterized emotions based on similarities of stimuli and biological processes, following the ideas that have been described already by Charles Darwin – clearly an attempt that focuses on observations and evolution.

One of Tomkins' mentees, Paul Ekman, put in question the existing emotion theories that proclaimed that facial expressions of emotion are socially learned and therefore vary from culture to culture. Ekman, Sorenson and Friesen challenged this view[28] in a field study with the outcome that facial displays of fundamental emotions are not learned but innate.

[25] Darwin 1872.
[26] Gendron / Feldman Barrett 2009.
[27] Tomkins 1962.
[28] Ekman et al. 1969, passim.

However, there are culture-specific prescriptions about how and in which situations emotions are displayed. Based on the observation of facial behavior in early development or social interaction, Ekman's theory also postulates that emotions should be considered *discrete categories*[29] rather than continuous. Though this view allows for conceiving of emotions as having different intensities, it does not allow emotions to blend and leaves no room for more complex affective states in which individuals report the *co-occurrence of like-valenced discrete emotions*.[30] (This and other theory postulates were widely criticized and disputed in literature.[31])

Ekman and colleagues, however, defined clearly how basic emotions can be distinguished from other emotions: There are distinctive universal signals, the presence in other primates, distinctive phyiosology, distinctive universals in antecedent events, coherence in the emotional response, a quick onset, a brief duration, an automatic appraisal, and an automatic, unbidden occurrence. The set of emotions that is typically referred to as "Ekman emotions" consists of anger, fear, joy, sadness, surprise, and disgust. Given that this set of emotions is relevant for many studies, and that these emotion categories do not deserve further explanation to most people, it constitutes a popular basis for computational analysis.

## 2.2 Plutchik's Wheel of Emotions

Another influential model of emotions was proposed by Robert Plutchik in the early 1980s.[32] The important difference between Plutchik's theory and Ekman's theory is that apart from a small set of basic emotions, all other emotions are mixed and derived from the various combinations of basic ones. He further categorized these other emotions into the *primary dyads* (very likely to co-occur), *secondary dyads* (less likely to co-occur) and *tertiary dyads* (seldom co-occur).

In order to represent the organization and properties of emotions as defined by his theory, Plutchik proposed a structural model of emotions known nowadays as *Plutchik's wheel of emotions*. The wheel (Abb.) is constructed in the fashion of a color wheel, with similar emotions placed closer together and opposite emotions 180 degrees apart. The intensity of an emotion in the wheel depends on how far from the center a part of a petal is, i.e., emotions become less distinguishable the further they are from the center of the wheel. Essentially, the wheel is constructed from eight basic bipolar emotions: *joy* versus *sadness*, *anger* versus *fear*, *trust* versus *disgust*, and *surprise* versus *anticipation*. The blank spaces between the leaves are so-called *primary dyads* – emotions that are mixtures of two of the primary emotions.

The *wheel model of emotions* proposed by Plutchik had a great impact on the field of affective computing being primarily used as a basis for emotion categorization in emotion recognition from text.[33] However, some postulates of the theory are criticized, for example, there is no empirical support for the wheel structure[34]. Another criticism is that Plutchik's model of emotion does not explain the mechanisms by which emotions emerge from the *basic* emotions, nor does it provide reliable measurements of these emotions.[35]

---

[29] Ekman 1993, passim.
[30] Barrett 1998.
[31] Russell 1994; Russell et al. 2003; Gendron et al. 2014; Barrett 2017.
[32] Plutchik 1991.

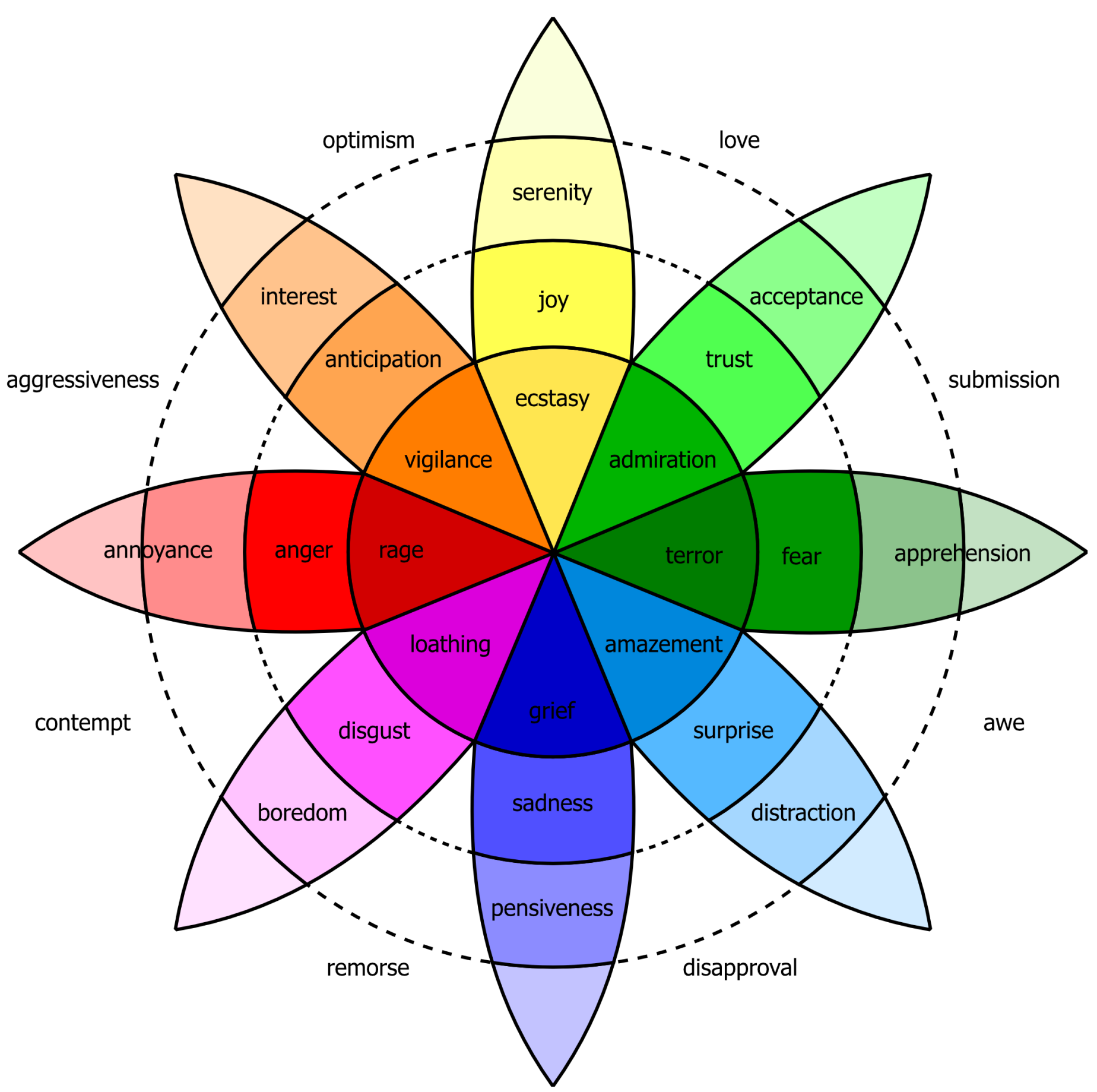

Fig. 1: Plutchik's wheel of emotions. In: Wikipedia, the free Encyclopedia: Robert Plutchik. Article from 20.09.2019. [online] [Plutchik 2011. PD.]

## 2.3 Russel's Circumplex Model

Attempts to overcome the shortcomings of basic emotion theories and its unfitness for clinical studies led researchers to suggest various dimensional models, the most prominent of which is the circumplex model of affect proposed by James Russel.[36] The word *circumplex* in the name of the model refers to the fact that emotional episodes do not cluster at the axes but rather at the periphery of a circle (Abb.). At the core of the circumplex model is the notion of two dimensions plotted on a circle along horizontal and vertical axes. These dimensions are *valence* (how pleasant or unpleasant one feels) and *arousal* (the degree of calmness or excitement). The number of dimensions is not strictly fixed and there are adaptations of the model that incorporate more dimensions. One example of this is the *Valence-Arousal-Dominance model* that adds an additional dimension of dominance, the degree of control one feels over the situation that causes an emotion.[37]

---

[33] Cambria et al. 2012, passim; Kim et al. 2012, passim; Suttles / Ide 2013, passim; Borth et al. 2013, passim; Abdul-Mageed / Ungar 2017, passim.
[34] Smith / Schneider 2009, passim.
[35] Richins 1997, passim.
[36] Russell 1980.

By moving from discrete categories to a dimensional representation, the researchers are able to account for subjective experiences that do not fit nicely into the isolated non-overlapping categories. Accordingly, each affective experience can be depicted as a point in a *circumplex* that is described by only two parameters – *valence* and *arousal* – without need for labeling or reference to emotion concepts for which a name might only exist in particular subcommunities or which are difficult to describe.[38] However, the strengths of the model turned out to be its weaknesses: for example, it is not clear whether there are basic dimensions in the model[39] nor is it clear what should be done with qualitatively different events of *fear*, *anger*, *embarrassment* and *disgust* that fall in identical places in the circumplex structure.[40] Despite these shortcomings, the circumplex model of affect is popular in psychologic and psycholinguistic studies, because both dimensions can reliably be measured.[41] In computational linguistics, the circumplex model is applied when the interest is in continuous measurements of *valence* and *arousal* rather than in the specific discrete emotional categories.

There are other models which locate discrete emotion categories in a dimensional space, however, these have not been used in computational literary studies. These theories are based on appraisal theories[42] and state that different dimensions, which measure how a stimulus event is cognitively evaluated enable different sets of emotions. The work by Smith/Ellsworth[43] shows that the six dimensions of (1) how pleasant an event is, (2) how much effort an event can be expected to cause, (3) how certain the experiencer is in a specific situation, (4) how much attention is devoted to the event, (5) how much responsibility the experiencer of the emotion holds for what has happened, and (6) how much the experiencer has control over the situation, explain 15 discrete emotions.

[37] Bradley / Lang 1994.
[38] Russell 2003.
[39] Larsen / Diener 1992.
[40] Russell / Barrett 1999.
[41] Mauss / Robinson 2009.
[42] Scherer 2005.
[43] Smith / Ellsworth (1985)

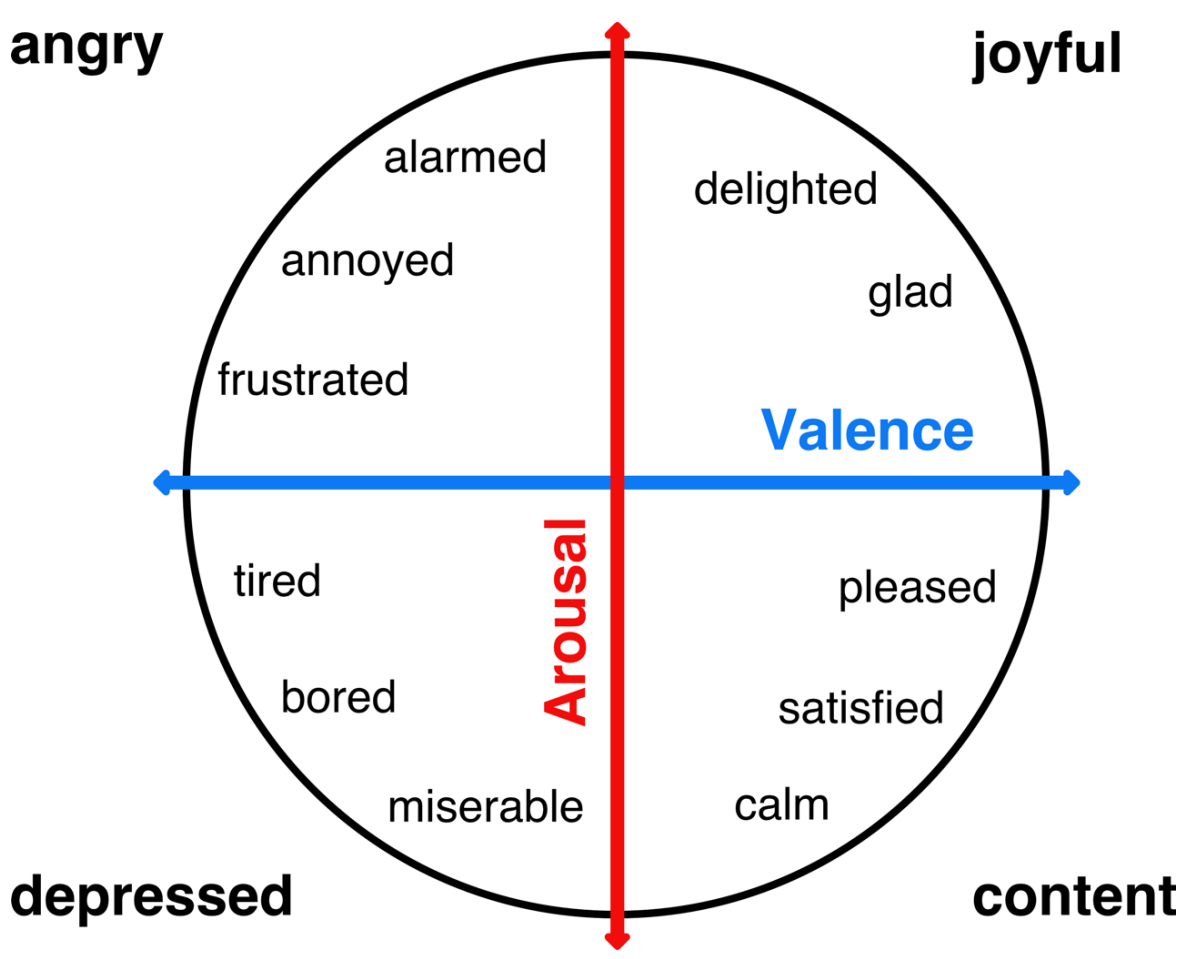

Fig. 2: Circumplex model of affect: Horizontal axis represents the valence dimension, the vertical axis represents the arousal dimension. (Drawn after Posner et al. 2005)

## 3 Emotion Analysis in Non-computational Literary Studies

In the past, literary and art theories often disregarded the importance of the aesthetic and affective dimension of literature, which in part stemmed from the rejection of old-fashioned literary history that had explained the meaning of art works by the biography of the author.[44] However, the affective turn taken by a wide range of disciplines in the past two decades – from political and sociological sciences to neurosciences or media studies – has refueled the interest of literary critics in human affects and sentiments.

We said in Section 1 that there seems to be a consensus among literary critics that literary art and emotions go hand in hand. However, one might be challenged to define the specific way in which emotions come into play in the text. The exploration of this problem is presented by van Meel.[45] Underpinning the centrality of human destiny, hopes, and feelings in the themes of many artworks – from painting to literature – van Meel explores how emotions are involved in the production of arts. Pointing out big differences between the two media in their attempts to depict human emotions (painting conveys nonverbal behavior directly, but lacks temporal dimensions that novels have and use to describe emotions), van Meel provides an analysis of the nonverbal descriptions used by the writers to convey their characters' emotional behavior. Description of visual characteristics, van Meel speculates, responds to a fundamental need of a reader to build an image of a person and their behavior. Moreover, nonverbal descriptions add important information that can in some cases play a crucial hermeneutical role, such as in Kafka's *Der Prozess*, where the fatal decisions for K. are made clear by gestures rather than words. His verdict is not announced, but is implied by the judge who refuses a handshake. The same applies to his death sentence that is conveyed to him by his executioners playing with a butcher's knife above his head. These aspects how emotions are communicated clearly point to challenges for computational methods – implicit descriptions, world knowledge, and inference steps that are grounded in combinations of text and readers' experiences have not been tackled with computational methods yet.

---

[44] Sætre et al. 2014b.
[45] van Meel 1995.

A hermeneutic approach through the lense of emotions is presented by Kuivalainen[46] and provides a detailed analysis of linguistic features that contribute to the characters' emotional involvement in Mansfield's prose. The study shows how, through the extensive use of adjectives, adverbs, deictic markers, and orthography, Mansfield steers the reader towards the protagonist's climax. Subtly shifting between psycho-narration and free indirect discourse, Mansfield is making use of evaluative and emotive descriptors in psycho-narrative sections, often marking the internal discourse with dashes, exclamation marks, intensifiers, and repetition that thus trigger an emotional climax. Various deictic features introduced in the text are used to pinpoint the source of emotions, which helps in creating a picture of characters' emotional world. Verbs (especially in the present tense), adjectives, and adverbs serve the same goal in Mansfield's prose of describing the characters' emotional world. Going back and forth from psycho-narration to free indirect discourse provides Mansfield with a tool to point out the significant moments in the protagonists' lives and establish a separation between characters and narration. This study illustrates another challenge for automatic methods. Computational models mostly rely on isolated, comparable short, units of the text. The broader context, let alone the development of characters, are mostly ignored in computational analysis – a prediction depends on the local description and is not conditioned on previous experiences. That is a clear disadvantage of distant reading methods to close reading.

Both van Meel's and Kuivalainen's works, separated from each other by more than a decade, underpin the importance of emotions in the interpretation of characters' traits, hopes, and tragedy. Other authors find these connections as well. For example, Barton[47] proposes instructional approaches to teach school-level readers to interpret character's emotions and use this information for story interpretation. Van Horn[48] shows that understanding characters emotionally or trying to help them with their problems made reading and writing more meaningful for middle school students.

Emotions in text are often conveyed with emotion-bearing words.[49] At the same time their role in the creation and depiction of emotion should not be overestimated. That is, saying that someone looked angry or fearful or sad, as well as directly expressing characters' emotions, are not the only ways authors build believable fictional spaces filled with characters, action, and emotions. In fact, many novelists strive to express emotions indirectly by way of figures of speech or catachresis,[50] first of all because emotional language can be ambiguous and vague, and, second, to avoid any allusions to Victorian emotionalism and pathos.

How can an author convey emotions indirectly? A book chapter by Hillis Miller in *Exploring Text and Emotions*[51] seeks the answer to exactly this question. Using Conrad's *Nostromo* opening scenes as material, Hillis Miller shows how Conrad's descriptions of an imaginary space generate emotions in readers without direct communication of emotions. Conrad's

---

[46] Kuivalainen 2009.
[47] Barton 1996.
[48] Van Horn 1997.
[49] Johnson-Laird / Oatley 1989.
[50] Hillis Miller 2014.
[51] Sætre et al. 2014a.

*Nostromo* opening chapter is an objective description of Sulaco, an imaginary land. The description is mainly topographical and includes occasional architectural metaphors, but it combines wide expanse with hermetically sealed enclosure, which generates depthless emotional detachment[52]. Through the use of present tense, Conrad makes the readers suggest that the whole scene is timeless and does not change. The topographical descriptions are given in a pure materialist way: there is nothing behind clouds, mountains, rocks, and sea that would matter to humankind, not a single feature of the landscape is personified, and not a single topographical shape is symbolic. Knowingly or unknowingly, Miller argues, by telling the readers what they should see – with no deviations from truth – Conrad employs a trope that perfectly matches Kant's *concept of the sublime*. Kant's view of poetry was that true poets tell the truth without interpretation; they do not deviate from what their eyes see. Conrad, or to be more specific, his narrator in *Nostromo*, is an example of sublime seeing with a latent presence of strong emotions. On the one hand, Conrad's descriptions are cool and detached. This coolness is caused by the indifference of the elements in the scene. On the other hand, by dehumanizing sea and sky, Conrad generates awe, fear, and a dark foreboding about the kinds of life stories that are likely to be enacted against such a backdrop[53].

Hillis Miller's analysis resonates with some premises from emotion theory that we have discussed previously, namely, Plutchik's belief that emotions should be studied not by a certain way of expression but by the overall behavior of a person. Considering that such a formula cannot be applied to all literary theory studies about emotions (as not all authors choose to convey emotions indirectly, as well as not all authors tend to comment on characters' nonverbal emotional behavior), it seems that one should search for a balance between low-level linguistic feature analysis of emotional language and a rigorous high-level hermeneutic inquiry dissecting the form of the novel and its under-covered philosophical layers.

We recommend the essay[54] by Katja Mellmann for further details on that topic.

## 4 Emotion and Sentiment Analysis in Computational Literary Studies

With this section, we proceed to an overview of the existing body of research on computational analysis of emotion and sentiment in computational literary studies. An overview of the papers including their properties is shown in Table 1. The table, as well as this section, is divided into several subsections, each of which corresponds to a specific application of emotion analysis to literature. Section 4.1 reviews the papers that deal with the classification of literary texts in terms of emotions they convey; Section 4.2 examines the papers that address text classification by genre or other story-types based on sentiment and emotion features; Section 4.3 is dedicated to research in modeling sentiments and emotions in texts from previous centuries, as well as research dealing with applications of sentiment analysis to texts written in the past; Section 4.4 provides an overview of sentiment analysis applications to character analysis and character network construction, and Section 4.5 is dedicated to more general applications.

[52] Hillis Miller 2014, p. 93.
[53] Hillis Miller 2014, p. 115.
[54] Mellmann 2002

## 4.1 Emotion Classification

A straightforward approach to emotion analysis is text classification[55]. Indeed, emotion classification is one of the most popular subtasks and finds application in several downstream tasks. A fundamental question of such a classification is how to find the best input representations and algorithms to classify the data (sentences, paragraphs, entire documents) into predefined classes. When applied to literature, such a classification may be of use for grouping different literary texts in digital collections based on the emotional properties of the stories or to perform other analyses regarding the distribution of emotions in subcollections. For example, books or poems can be grouped based on the emotions they convey or based on whether or not they have happy endings or not.

### 4.1.1 Classification based on emotions

Barros et al.[56] aim at answering two research questions: 1) is the classification of Quevedo's works proposed by the literary scholars consistent with the sentiment reflected by the corresponding poems; and 2) which learning algorithms are the best for the classification (the latter being an engineering question that is inherent in many of the papers that we discuss)? They perform a set of experiments on the classification of 185 Francisco de Quevedo's poems that are divided by literary scholars into four categories and that Barros et al. map to emotions. Using the terms *joy*, *anger*, *fear*, and *sadness* as points of reference, Barros et al. construct a list of emotion words by looking up the synonyms of English emotion words and adjectives associated with these four emotions and translating them into Spanish. This leads to a novel and task-specific lexicon, to which each poem is then compared, based on normalized term counts. The experiments show the superiority of decision trees as classification approach which can further be improved by rebalancing the collection via resampling. Based on these results the authors conclude that a meaningful classification of the literary pieces based only on the emotion information is possible.

A more modern corpus selection of poetry is the object of analysis by Reed[57]. The authors offer a *proof-of-concept* for performing sentiment analysis on twentieth-century American poetry with dictionary-based black-box sentiment analysis systems that output the polarity of a text. Specifically, they analyze the expression of emotions in the poetry of the *Black Arts Movement* of the 1960s and 1970s. The goal of the project is to understand how the feelings associated with injustice are coded in terms of race and gender, and what sentiment analysis can show us about the relations between affect and gender in poetry. Reed notes that the surface affective value of the words does not always align with their more nuanced affective meaning shaped by poetic, social, and political contexts. Therefore, this study can be seen

---

[55] Liu 2015, p. 47.
[56] Barros et al. 2013.
[57] Reed 2018.

as a critical reflection on methodological choices.

Yu58 look closer at linguistic patterns that characterize the genre of sentimentalism in early American novels. They analyze five novels from the mid-nineteenth century and annotate the emotionality of each of the chapters as *high* or *low (not: positive or negative!)*. This approach is noteworthy, as the unit of analysis is comparably large in contrast to most sentiment analysis methods. Each chapter is classified with standard configurations of support vector machines and naïve Bayes classifiers, as highly emotional or the opposite. The results of the evaluation suggest that arbitrary feature reduction steps such as stemming and stopword removal should be taken very carefully, as they may affect the prediction.

Volkova[59] did not focus on the classification of emotions automatically, but tackles the task of annotation in more detail. The authors observe that annotation of literature, in their concrete case of fairy tales, is challenging, and that it is hard to obtain an acceptable agreement. An interesting innovative element in this study is that annotators were not presented a predefined unit to annotate – they were allowed to decide by themselves which granularity is most reasonable. That is different to the other studies mentioned before in this section. Further, a main finding was that short instances lead to a lower agreement.

Finally, an interesting study by Ashok et al. did not classify emotions regarding a variable motivated by literary studies. They use sentiment polarity as one component to predict the success of a book. While such studies (similarly the prediction of citation counts, etc.) are often criticized, the authors present some interesting, but also perhaps non-surprising finding: unsuccessful stories contain more discriminative words that have a negative connotation.

### 4.1.2 Classification of happy ending vs. non-happy endings

A particular use case of emotion classification is to look closer at particular parts of a text. Zehe et al.[60] argue that automatically recognizing a happy ending as a major plot element could help to better understand a plot structure as a whole. To show that this is possible, they classify 212 German novels written between 1750 and 1920 as having happy or non-happy endings. A novel is considered to have a happy ending if the situation of the main characters in the novel improves towards the end or is constantly favorable. The novels were manually annotated with this information by domain experts. For feature extraction, the authors first split each novel into *n* segments of the same length. They then calculate sentiment values for each of the segments based on a normalized word frequency with a German version of the *NRC Word-Emotion Association Lexicon*[61]. An automatic classification with support vector machines achieves reasonable and encouraging results.

---

[58] Yu 2008.
[59] Volkova et al. 2010.
[60] Zehe et al. 2016.
[61] Mohammad / Turney 2013.

## 4.2 Genre and Story-type Classification

The papers we have discussed so far focus on understanding the emotion associated with units of texts. This extracted information can further be used for downstream tasks and also for downstream evaluations. We discuss the following downstream classification cases here. The papers in this category use sentiment and emotion features for a higher-level classification, namely story-type clustering and literary genre classification. The assumption behind these works is that different types of literary text may show different composition and distribution of emotion vocabulary and thus can be classified based on this information. The hypothesis that different literary genres convey different emotions stems from common knowledge: we know that horror stories instill *fear* and that mysteries evoke *anticipation* and *anger* while romances are filled with *joy* and *love*. However, as we will see in this section, the task of automatic classification of these genres is not always that straightforward and reliable.

### 4.2.1 Story-type clustering

Similarly to Zehe et al., Reagan et al.[62] are interested in automatically understanding a plot structure as a whole, but not limited to a book ending. The inspiration for their work comes from Kurt Vonnegut's lecture on emotional arcs of stories.[63] Reagan et al. test the idea that the plot of each story can be visualized as an *emotional arc*, i.e., a time series graph, where the *x*-axis represents a time point in a story, and the *y*-axis represents the events happening to the main characters that can be favorable (peaks on a graph) or unfavorable (troughs on a graph). As Vonnegut puts it, the stories can be grouped by these *arcs* and the number of such groupings is limited. To test this idea, Reagan et al. collect the 1,327 most popular books from the Project Gutenberg.[64] Each book is then split into segments for which happiness scores are calculated and compared. The results of the analysis show support for six emotional patterns that are shared between subgroupings of the corpus. Additionally, Reagan et al. find that some patterns are more popular among readers, based on download counts, than others.

### 4.2.2 Genre classification

There are other studies[65] that are similar in spirit to the work done by Reagan. Samothrakis and Fasli examine the hypothesis that different genres clearly have different emotion patterns to reliably classify them with machine learning. To that end, they collect works of the genres *mystery*, *humor*, *fantasy*, *horror*, *science fiction* and *western* from the Project Gutenberg. Using WordNet-Affect66 to detect emotion words as categorized by Ekman's fundamental emotion classes, they calculate an emotion score for each sentence in the text.

---

[62] Reagan et al. 2016.
[63] Vonnegut 2010 (2005).
[64] Project Gutenberg.
[65] Samothrakis / Fasli 2015; Kim et al. 2017a; Kim et al. 2017b.
[66] Strapparava / Valitutti 2004.

Each work is then transformed into six vectors, one for each basic emotion. With a random forrest classifier, they show that genre classification is possible based on this information with performance scores significantly above average.

The study by Kim et al.[67] originates from the same premise as the work by Samothrakis and Fasli but puts emphasis on finding genre-specific correlations of emotion developments. They therefore link the motivation of Reagan with the one by Samothrakis and Fasli. Extending the set of tracked emotions to Plutchik's classification, Kim et al. collect 2,000 books from the Project Gutenberg that belong to five genres found in the Brown corpus[68], namely *adventure*, *science fiction*, *mystery*, *humor* and *romance*. The authors extend the set of classification algorithms beyond random forests using a *multi-layer perceptron* and *convolutional neural networks*, which achieves the best performance. To understand how uniform the emotion patterns in different genres are, the authors introduce the notion of *prototypicality*, which is computed as average of all emotion scores. Using this as a point of reference for each genre Kim et al. use Spearman correlation to calculate the uniformity of emotions per genre. The results of this analysis suggest that *fear* and *anger* are the most salient plot devices in fiction, while *joy* is only of mediocre stability, which is in line with findings of Samothrakis and Fasli.

The study by Henny-Khramer[69] pursues two goals: 1), to test whether different subgenres of Spanish American literature differ in degree and kind of emotionality, and 2), whether emotions in the novels are expressed in direct speech of characters or in narrated text. To that end, they conduct a subgenre classification experiment on a corpus of Spanish American novels using sentiment values as features. To answer the first question, each novel is split into five segments and for each sentence in the segment the emotion score (polarity values + Plutchik's basic emotions) is calculated using SentiWordNet[70] and NRC[71] dictionaries. The analysis of feature importance shows that the most salient features come from the sentiment scores calculated from the characters' direct speech and that novels with higher values of positive speech are more likely to be sentimental novels. This is an interesting variant of the beforehand mentioned studies – it is important to distinguish characters' speech from other parts of the text.

There are some limitations to the studies presented in this section. On the one hand, it is questionable how reliable *coarse emotion scoring* is that takes into account only presence or absence of words found in specialized dictionaries and overlooks negations and modifiers that can either negate an emotion word or increase/decrease its intensity. On the other hand, a limited view of the emotional content as a sum of emotion bearing words reserves no room for qualitative interpretation of the texts – it is not clear how one can distinguish between emotion words used by the author to express their sentiment, between words used to describe characters' feelings, and emotion words that characters use to address or describe other characters in a story.

---

[67] Kim et al. 2017a.
[68] Francis / Kucera 1979.
[69] Henny-Krahmer 2018.
[70] Baccianella et al. 2010.
[71] Mohammad / Turney 2013.

### 4.3 Structural Changes of Sentiment

The papers that we have reviewed so far approach the problem of sentiment and emotion analysis as a classification task. However, applications of sentiment analysis are not only limited to classification. In other fields, for example computational social sciences, sentiment analysis can be used for analyzing political preferences of the electorate or for mining opinions about different products or topics. Similarly, several digital humanities studies incorporate sentiment analysis methods in a task of mining sentiments and emotions of people who lived in the past. The goal of these studies is not only to recognize sentiments, but also to understand how they were formed.

#### 4.3.1 Topography of emotions

Heuser et al.[72] start with a premise that emotions occur at a specific moment in time and space, thus making it possible to link emotions to specific geographical locations. Consequently, having such information at hand, one can understand which emotions are hidden behind certain landmarks. As a *proof-of-concept*, Heuser et al. build an interactive map of emotions in Victorian London[73] where each location is tagged with emotion labels. The underlying corpus for their analysis consists of English books from the eighteenth and nineteenth century, from which they extract frequently mentioned geographical locations of London. The presegmented data is then given to annotators who are asked to define whether each of the passages expressed *happiness* or *fear*, or *neutrality*. The same data is further analyzed with a dictionary-based sentiment classifier.

Some striking observations are made with regard to the data analysis. First, there is a clear discrepancy between fiction and reality – while toponyms from the West End with Westminster and the City are over-represented in the books, the same does not hold true for the East End with Tower Hamlets, Southwark, and Hackney. Hence, there is less information about emotions pertaining to these particular London locations. Another striking detail is that the resulting map is dominated by the neutral emotion. Heuser et al. argue that this has nothing to do with the absence of emotions but rather stems from the fact that emotions tend to be silenced in public domain, which influenced the annotators decision.

The space and time context are also used by Bruggman and Fabrikant[74] who model sentiments of Swiss historians towards places in Switzerland in different historical periods. As the authors note, it is unlikely that a historian will directly express attitudes towards certain toponyms, but it is very likely that words they use to describe those can bear some negative connotation (e.g., cholera, death). Correspondingly, such places should be identified as bearing negative sentiment by a sentiment analysis tool. Additionally, they study the changes of sentiment towards a particular place over time. Using the *General Inquirer* (GI) lexicon[75] to identify positive and negative terms in the document, they assign sentiment scores and conclude that the results of their analysis look promising, especially regarding negatively scored articles.

---

[72] Heuser et al. 2016.
[73] Interactive map of emotions in Victorian London.
[74] Bruggmann / Fabrikant 2014.
[75] Stone et al. 1968.

### 4.3.2 Tracking sentiment

Other papers in this category link sentiment and emotion to certain groups, rather than geographical locations. The goal of these studies is to understand how sentiment within and towards these groups was formed.

Taboada et al.[76] aim at tracking the literary reputation of six authors writing in the first half of the twentieth century. The research questions raised in the project are how the reputation is made or lost, and how to find correlation between what is written about the author and their work to the author's reputation and subsequent canonicity. The project's goal is to examine critical reviews of six authors' writing and to map information contained in texts critical to the author's reputation. The material they work with includes not only reviews, but also press notes, press articles, and letters to editors (including from the authors themselves). They collected and scanned 330 documents and tagged them with polarity words with custom-made sentiment dictionaries. The sentiment orientation of rhetorically important parts of the texts is then measured.

Chen et al.[77] aim to understand personal narratives of Korean *comfort women* who had been forced into sexual slavery by Japanese military during World War II. Adapting the *WordNet-Affect* lexicon,[78] Chen et al. build their own emotion dictionary to spot keywords in women's stories and map the sentences to emotion categories. By adding variables of time and space, Chen et al. provide a unified framework of collective remembering of this historical event as witnessed by the victims.

An interesting methodological contribution has been made by Gao et al.[79] Instead of using raw counts of polarity words over time, they propose that filters are used to smooth the time series, which further allows for other downstream applications.

Finally, an interesting project to follow is the Oceanic Exchanges[80] project[81] that started in late 2017. One goal of the project is to trace information exchange in nineteenth-century newspapers and journals using sentiment as one of the variables under analysis.

---

[76] Taboada et al. 2006; Taboada et al. 2008.
[77] Chen et al. 2012.
[78] Strapparava / Valitutti 2004.
[79] Gao et al. 2016
[80] Oceanic Exchanges.
[81] Cordell et al. 2017.

### 4.3.3 Sentiment recognition in historical texts

Other papers put emphasis not so much on the sentiments expressed by writers but instead focus on the particularities of historical language. Marchetti et al.[82] and Sprugnoli et al.[83] present the integration of sentiment analysis in the ALCIDE (Analysis of Language and Content in a Digital Environment) project[84]. The sentiment analysis module is based on *WordNet-Affect*, *SentiWordNet*[85] and *MultiWordNet*.[86] Each document is assigned a normalized polarity score. The overall conclusion of their work is that the assignment of a polarity in the historical domain is a challenging task largely due to lack of agreement on polarity of historical sources between human annotators.

Challenged by the problem of applicability of existing emotion lexicons to historical texts, Buechel et al.[87] propose a new method of constructing affective lexicons that would adapt well to German texts written up to three centuries ago. In their study, Buechel et al. use the representation of affect based on the *Valence-Arousal-Dominance model* (an adaptation of Russel's circumplex model, see Section 2.3). Presumably, such a representation provides a finer-grained insight into the literary text[88], which is more expressive than discrete categories, as it quantifies the emotion along three different dimensions. As a basis for the analysis, they collect German texts from the Deutsches Textarchiv89 written between 1690 and 1899. The corpus is split into seven slices, each spanning 30 years. For each slice they compute word similarities and obtain seven distinct emotion lexicons, each corresponding to specific time period. This allows for, the authors argue, the tracing of the shift in emotion association of words over time.

Finally, Leemans et al.[90] aim to trace historical changes in emotion expressions and to develop methods to trace these changes in a corpus of 29 Dutch language theatre plays written between 1600 and 1800. Expanding the Dutch version of *Linguistic Inquiry and Word Count* (LIWC) dictionary[91] with historical terms, the authors are able to increase the recall of emotion recognition with a dictionary. In addition, they develop a fine-grained vocabulary mapping body terms to emotions, and show that a combination of LIWC and their lexicon leads to improvement in the emotion recognition.

## 4.4 Character Network Analysis and Relationship Extraction

The papers reviewed above address sentiment analysis of literary texts mainly on a document level. This abstraction is warranted if the goal is to get an insight into the distribution of emotions in a corpus of books. However, emotions depicted in books do not exist in isolation but are associated with characters who are at the core of any literary narrative.[92] This leads us to ask what sentiment and emotion analysis can tell us about the characters. How emotional are they? And what role do emotions play in their interaction?

[82] Marchetti et al. 2014.
[83] Sprugnoli et al. 2016.
[84] .ALCIDE.
[85] Baccianella et al. 2010.
[86] Pianta et al. 2002.
[87] Buechel et al. 2017.
[88] Buechel et al. 2016.
[89] DTA.
[90] Leemans et al. 2017.
[91] Pennebaker et al. 2007.

Character relationships have been analyzed in computational linguistics from a graph theoretic perspective, particularly using social network analysis.[93] Fewer works, however, address the problem of modeling character relationships in terms of sentiment. Below we provide an overview of several papers that propose the methodology for extracting this information.

### 4.4.1 Sentiment dynamics between characters

Several studies present automatic methods for analyzing sentiment dynamics between plays' characters. The goal of the study by Nalisnick and Baird[94] is to track the emotional trajectories of interpersonal relationships. The structured format of a dialog allows them to identify who is speaking to whom, which makes it possible to mine character-to-character sentiment by summing the valence values of words that appear in the continuous direct speech and are found in the lexicon[95] of affective norms. The extension[96] of the previous research from the same authors introduces the concept of a sentiment network, a dynamic social network of characters. Changing polarities between characters are modeled as edge weights in the network. Motivated by the desire to explain such networks in terms of a general sociological model, Nalisnick and Baird test whether Shakespeare's plays obey the *Structural Balance Theory* by Marvel et al.[97] that postulates that a friend of a friend is also your friend. Using the procedure proposed by Marvel et al. on their Shakespearean sentiment networks, Nalisnick and Baird test whether they can predict how a play's characters will split into factions using only information about the state of the sentiment network after Act II. The results of their analysis are varied and do not provide adequate support for the Structural Balance Theory as a benchmark for network analysis in Shakespeare's plays. One reason for that, as the authors state, is inadequacy of their shallow sentiment analysis methods that cannot detect such elements of speech as irony and deceit that play a pivotal role in many literary works.

### 4.4.2 Character analysis and character relationships

Elsner[98] aims at answering the question of how to represent a plot structure for summarization and generation tools. To that end, Elsner presents a *kernel* for comparing novelistic plots at the level of character interactions and their relationships. Using sentiment as one of the characteristics of a character, Elsner demonstrates that the kernel approach leads to meaningful plot representation that can be used for a higher-level processing.

---

[92] Ingermanson / Economy 2009, p. 107.
[93] Agarwal et al. 2013; Elson et al. 2010.
[94] Nalisnick / Baird 2013a.
[95] Nielsen 2011.
[96] Nalisnick / Baird 2013b.
[97] Marvel et al. 2011.
[98] Elsner 2012; Elsner 2015.

Kim and Klinger[99] aim at understanding the causes of emotions experienced by literary characters. To that end, they contribute the REMAN corpus [100] of literary texts with annotations of emotions, experiencers, causes and targets of the emotions. The goal of the project is to enable the automatic extraction of emotions and causes of emotions experienced by the characters. The authors suggest that the results of coarse-grained emotion classification in literary text are not readily interpretable as they do not tell much about who the experiencer of the emotion is. Indeed, if a text mentions two characters, one of whom is *angry* and another one who is *scared* because of that, text classification models will only tell us that the text is about *anger* and *fear*. Hence, a finer-grained approach towards character relationship extraction is warranted. Kim and Klinger conduct experiments on the annotated dataset showing that the fine-grained approach to emotion prediction with long short-term memory networks outperforms *bag-of-words models*. At the same time, the results of their experiments suggest that joint prediction of emotions and experiencers can be more beneficial than studying these categories separately.

A tool presented by Jhavar and Mirza[101] provides a similar functionality: given an input of two character names from the *Harry Potter* series, the EMOFIEL[102] tool identifies the emotion flow between a given directed pair of story characters. These emotions are identified using categorical[103] and continuous[104] emotion models.

Egloff et al.[105] present an ongoing work on the *Ontology of Literary Characters* (OLC) that allows us to capture and infer characters' psychological traits from their linguistic descriptions. The OLC incorporates the *Ontology of Emotion*[106] that is based on both Plutchik's and Hourglass's[107] models of emotions. The ontology encodes 32 emotion concepts. Based on their natural language description, characters are attributed to a psychological profile along the classes of Openness to *experience*, *Conscientiousness*, *Extraversion*, *Agreeableness*, and *Neuroticism*. The ontology links each of these profiles to one or more archetypal categories of *hero*, *anti-hero*, and *villain*. Egloff et al. argue that, by using the semantic connections of the OLC, it is possible to infer the characters' psychological profiles and the role they play in the plot.

Kim and Klinger[108] propose the task of emotion relationship classification between fictional characters. They argue that joining character network analysis with sentiment and emotion analysis may contribute to a computational understanding of narrative structures, as characters are at the center of any plot development. Building a corpus of 19 fan fiction short stories and annotating it with emotions, Kim and Klinger propose several models to classify emotion relations of characters. They show that a deep learning architecture with character position indicators is the best for the task of predicting both directed and undirected emotion relations in the associated social network graph. As an extension to this study, Kim and Klinger[109] explore how emotions are expressed between characters in the same corpus via various non-verbal communication channels.[110] They find that facial expressions are predominantly associated with *joy* while gestures and body postures are more likely to occur with *trust*.

---

[99] Kim / Klinger 2018.
[100] Reman.
[101] Jhavar / Mirza 2018.
[102] Emofiel.
[103] Plutchik 1991.
[104] Russell 1980.
[105] Egloff et al. 2018.
[106] Patti et al. 2015.
[107] Cambria et al. 2012.
[108] Kim / Klinger 2019b.

Finally, a small body of work focuses on mathematical modeling of character relationships. Rinaldi et al.[111] contribute a model that describes the love story between the Beauty and the Beast through ordinary differential equations. Zhuravlev et al.[112] introduce a distance function to model the relationship between the protagonist and other characters in two masochistic short novels by Ivan Turgenev and Sacher-Masoch. Borrowing some instruments from the literary criticism and using ordinary differential equations, Zhuravlev et al. are able to reproduce the temporal and spatial dynamics of the love plot in the two novellas more precisely than it had been done in previous research. Jafari et al.[113] present a dynamic model describing the development of character relationships based on differential equations. The proposed model is enriched with complex variables that can represent complex emotions such as coexisting *love* and *hate*.

## 4.5 Other Types of Emotion Analysis

We have seen that sentiment analysis as applied to literature can be used for a number of downstream tasks, such as classification of texts based on the emotions they convey, genre classification based on emotions, and sentiment analysis in the historical domain. However, the application of sentiment analysis is not limited to these tasks. In this concluding part of the survey, we review some papers that do not formulate their approach to sentiment analysis as a downstream task. Often, the goal of these works is to understand how sentiments and emotions are represented in literary texts in general, and how sentiment or emotion content varies across specific documents or a collection of them with time, where time can be either relative to the text in question (from beginning to end) or to the historical changes in language (from past to present). Such information is valuable for gaining a deeper insight into how sentiments and emotions change over time, allowing us to bring forward new theories or shed more light onto existing literary or sociological theories.

### 4.5.1 Emotion flow analysis and visualization

A set of authors aimed to visualize the change of emotion content through texts or across time. One of the earliest works in this direction is a paper by Anderson and McMaster[114] that starts from the premise that reading enjoyment stems from the affective tones of a text. These affective tones create a conflict that can rise to a climax through a series of crises, which is necessary for a work of fiction to be attractive to the reader. Using a list of 1,000 of the most common English words annotated with valence, arousal, and dominance ratings,[115] they calculate the conflict score by taking the mean of the ratings for each word in a text passage. The more negative the score is, the higher the conflict is, and vice versa. Additionally, they plot conflict scores for each consecutive 100 words of a test story and provide qualitative analysis of the peaks. They argue that a reader who has access to the text would be able to find correlation between events in the story and peaks on the graph. However, the authors still stress that such interpretation remains dependent upon the judgement of the reader. Further, other contributions by the authors are based on the same premises.[116]

---

[109] Kim / Klinger 2019a.
[110] Their analysis is based on van Meel 1995 we mentioned in Section 3.
[111] Rinaldi et al. 2013.
[112] Zhuravlev et al. 2014.
[113] Jafari et al. 2016.
[114] Anderson / McMaster 1986.

Alm and Sproat[117] present the results of the emotion annotation task of 22 tales by the Grimm brothers and evaluate patterns of emotional story development. They split emotions into *positive* and *negative* categories and divide each story into five parts from which aggregate frequency counts of combined emotion categories are computed. The resulting numbers are plotted on a graph that shows a wave-shaped pattern. From this graph, Alm and Sproat argue, one can see that the first part of the fairy tales is the least emotional, which is probably due to scene setting, while the last part shows an increase in positive emotions, which may signify the happy ending.

Two other studies by Mohammad[118] focus on differences in emotion word density as well as emotional trajectories between books of different genres. Emotion word density is defined as a number of times a reader will encounter an emotion word on reading every *X words*. In addition, each text is assigned several emotion scores for each emotion that are calculated as a ratio of words associated with one emotion to the total number of emotion words occurring in a text. Both metrics use the *NRC Affective Lexicon* to find occurrences of emotion words. They find that fairy tales have significantly higher *anticipation*, *disgust*, *joy* and *surprise* word densities, but lower *trust* word densities when compared to novels.

A work by Klinger et al.[119] is a case study in an automatic emotion analysis of Kafka's *Amerika* and *Das Schloss*. The goal of the work is to analyze the development of emotions in both texts as well as to provide a character-oriented emotion analysis that would reveal specific character traits in both texts. To that end, Klinger et al. develop German dictionaries of words associated with Ekman's fundamental emotions plus contempt and apply them to both texts in question to automatically detect emotion words. The results of their analysis for *Das Schloss* show a striking increase of *surprise* towards the end and a peak of *fear* shortly after start of chapter 3. In the case of *Amerika*, the analysis shows that there is a decrease in *enjoyment* after a peak in chapter 4.

A similar study by Schmitd and Burghardt[120] also works on German text – but focuses on the mostly neglected domain of theater plays, more concretely the plays by Lessing. They perform an annotation study and subsequently analyze different established emotion lexicons to recover the emotion automatically. The configuration of the best performing system shows the highest accuracy of 0.7, while a majority baseline obtains 0.695.

[115] Heise 1965.
[116] Anderson / McMaster 1982; Anderson / McMaster 1993.
[117] Alm / Sproat 2005.
[118] Mohammad 2011; Mohammad 2012.
[119] Klinger et al. 2016.
[120] Schmidt / Burghardt 2018

Yet another work that tracks the flow of emotions in a collection of texts is presented by Kim et al.[121] The authors hypothesize that literary genres can be linked to the development of emotions over the course of text. To test this, they collect more than 2,000 books from five genres (*adventure*, *science fiction*, *mystery*, *humor* and *romance*) from Project Gutenberg and identify prototypical emotion shapes for each genre. Each novel in the corpus is split into five consecutive equally-sized segments (following the five-act theory of dramatic acts).[122] All five genres show close correspondence with regard to *sadness*, *anger*, *fear* and *disgust*, i.e., a consistent increase of these emotions from Act 1 to Act 5, which may correspond to an entertaining narrative. *Mystery* and *science fiction* books show increase in *anger* towards the end, and *joy* shows an inverse decreasing pattern from Act 1 to Act 2, with the exception of *humor*.

The work by Kakkonen and Galic Kakkonen[123] aims at supporting the literary analysis of *Gothic* texts at the sentiment level. The authors introduce a system called *SentiProfiler* that generates visual representations of affective content in such texts and outlines similarities and differences between them, however, without considering the temporal dimension. The *SentiProfiler* uses *WordNet-Affect* to derive a list of emotion-bearing words that will be used for analysis. The resulting sentiment profiles for the books are used to visualize the presence of sentiment in a particular document and to compare two different texts.

### 4.5.2 Miscellaneous

In this section, we review studies that are different in goals and research questions from the papers presented in previous sections and do not constitute a category on their own.

Koolen[124] claims that there is a bias among readers that put works by female authors on par with »women's books«, which, as stated by the author, tend to be perceived as of lower literary quality. She investigates how much »women's books« (here, *romantic* novels written by women) differ from novels perceived as literary (female and male-authored literary fiction). The corpus used in the study is a collection of European and North-American novels translated into Dutch. Koolen uses a Dutch version of the *Linguistic Inquiry* and *Word Count*,[125] a dictionary that contains content and sentiment-related categories of words to count the number of words from different categories in each type of fiction. Her analysis shows that romantic novels contain more positive emotions and words pertaining to friendship than in literary fiction. However, female-authored literary novels and male-authored ones do not significantly differ on any category.

Kraicer and Piper[126] explore the women's place within contemporary fiction starting from the premise that there is a near ubiquitous underrepresentation and decentralization of women. As a part of their analysis, Kraicer and Piper use sentiment scores to look at social balance and »antagonism«, i.e., how different gender pairings influence positive and negative language surrounding the co-occurrence of characters (using the sentiment dictionary presented by Liu[127] to calculate a sentiment score for a character pair). Having analyzed a set of 26,450 characters from 1,333 novels published between 2001 and 2015, the authors find that sentiment scores give little indication that the character's gender has an effect on the state of social balance.

---

[121] Kim et al. 2017b.
[122] Freytag 1863.
[123] Kakkonen / Galic Kakkonen 2011.
[124] Koolen 2018.
[125] Boot et al. 2017.
126 Kraicer / Piper 2019.

Morin and Acerbi[128] focus on larger-scale data spanning a hundred thousand of books. The goal of their study is to understand how emotionality of written texts changed throughout the centuries. Having collected 307,527 books written between 1900 and 2000 from the Google Books corpus[129] they collect, for each year, the total number of case-insensitive occurrences of emotion terms that are found under positive and negative taxonomies of *LIWC* dictionary.[130] The main findings of their research show that emotionality (both *positive* and *negative* emotions) declines with time, and this decline is driven by the decrease in usage of positive vocabulary. Morin and Acerbi remind us that the *Romantic* period was dominated by emotionality in writing, which could be the effect of a group of writers who wrote above the mean. If one assumes that each new writer tends to copy the emotional style of their predecessors, then writers at one point of time are disproportionally influenced by this group of above-the-mean writers. However, this trend does not last forever and, sooner or later, the trend reverts to the mean, as each writer reverts to a normal level of emotionality.

An earlier work[131] written in collaboration with *Acerbi* provides a somewhat different approach and interpretation of the problem of the decline in positive vocabulary in English books of the twentieth century. Using the same dataset and lexical resources (plus *WordNet-Affect*) Bentley et al. find a strong correlation between expressed negative emotions and the *U.S. economic misery index*, which is especially strong for the books written during and after the World War I (1918), the Great Depression (1935), and the energy crisis (1975). However, in the present study,[132] the authors argue that the extent to which positive emotionality correlates with subjective well-being is a debatable issue. Morin and Acerbi provide more possible reasons for this effect as well as detailed statistical analysis of the data, so we refer the reader to the original paper for more information.

## 5 Discussion and Conclusion

We have shown throughout this survey that there is a growing interest in sentiment and emotion analysis within computational literary studies as one main field of digital humanities. Given the fact that DH have emerged into a thriving science within the past decade, it may safely be said that this direction of research is relatively new. It further constitutes an interesting field that connects literary studies and computational linguistics.

---

127 Liu et al. 2010.
128 Morin / Acerbi 2017.
129 Google Books Corpus.
130 Pennebaker et al. 2007.
131 Bentley et al. 2014.
132 Morin / Acerbi 2017.

In computational linguistics, sentiment analysis started more than two decades ago and is nowadays an established field that has dedicated workshops and tracks in the main conferences. Moreover, a recent meta-study by Mäntylä et al.[133] shows that the number of papers in sentiment analysis is rapidly increasing each year. Indeed, the topic has not yet outrun itself and we should not expect to see it vanishing within the next decade or two, provided that no significant paradigm shift in the computational sciences takes place. In addition, there are still many unsolved challenges. For each novel representation-learning approach, the question arises how sentiment concepts can be appropriętly represented. For many languages in the world the number of resources is low and it is not even known if established approaches could simply be transferred. To leverage these issues, research on multilingual methods that induce models in resource-scarce environments is an interesting modern direction, and a promising and rewarding field. All these developments on machine learning models, domain adaptation, pretraining and fine-tuning will also be beneficial for the digital humanities, but we cannot expect that all particular challenges that arise from research questions in literary studies will be solved in this field that focuses on generalizable methods.

Digital humanties has specific needs that cannot be readily addressed by existing methods or those that are developed in the future, in computational linguistics, machine learning, and computer science in general. As we have seen in this survey, most of the works rely on affective lexicons and word counts, a technique for detecting emotions in literary text first used by Anderson and McMaster in 1982.[134] Even the most recent works base the interpretation of the results on the use of dictionaries and counts of emotion-bearing words in a text, passage, or sentence. In fact, around 70% of the papers we discussed in Section 4 substantially rely on the use of various lexical resources for detecting emotions. We identify a set of particular challenges that hold for digital humanities and computational literary studies and that are presumable reasons for that choice.

*The object of research is the central element.* In contrast to computational linguistics, the goal of digital humanities is not to develop generalizable methods. The goal is, instead, to develop those methods that are helpful for a particular research question; and in contrast to computational linguistics, this includes tasks that only very few people work on. It would be a huge advantage if those methods generalized and could be reused, however, it is not a primary goal. Instead, an emotion analysis method for a particular scholar who analyzes texts from a particular subset, for instance genre, period, or author needs to work well for this subset. It might not be feasable to develop sophisticated deep learning methods for each of these approaches, just to be used once.

*Transparency of the computational method is not a bonus; it is a crucial property.* In digital humanities, research is often exploratory. The application of an existing method on a corpus can lead to new findings, but it is common that an interactive application of a method to explore a phenomenon is even more promising. Such interactive application requires full control by the user in real time – and that is something that pretrained deep neural methods cannot (yet) provide. However, emotion lexicons that point to particular aspects in the text in a transparent manner do, despite of their disadvantages.

[133] Mäntylä et al. 2018.
[134] Anderson / McMaster 1982.

*Computational expertise is not sufficient.* In computational research disciplines, a minimum amount of understanding of the respective domain is helpful but not necessarily (always) required. Particularly in recent years, with the development of end-to-end learning methods that hardly explain decisions, it became common to purely rely on performance measures (though this changes with recent research on explainable artificial intelligence). In contrast, in computational literary studies, knowledge of the domain is required. Without it, research questions cannot be answered. This is not a unique property of digital humanities as an interdisciplnary field. However, it is particularly challenging here, given its recent growth, fast development, and also the differences in the research culture between humanities and computer science (which are arguably smaller between, for instance, natural sciences and computer science, to which fields like computational chemistry or bioinformatics belong).

This leads to a set of challenges that need to be solved, while developing methods further. In contrast to most emotion analysis work in other domains (like social media or news), the unit of analysis should be larger. It is not sufficient to only analyze sentences in isolation (or even words). Instead, the overall development of characters, the story line as a whole need to be considered. This is a research direction that hardly received any attention; presumably because of technical challenges, but likely also due to the lack of annotated corpora that would be required to contain annotations on different levels. Further, these annotations need particular expertise from the annotators. It is not feasible to show an entire book to workers on a crowdsourcing platform to receive annotations on fine-grained levels (for characters and their developments). Therefore, for domains of interest, we point out that the development of corpora in computational literary studies are expected to be more expensive and take longer than in other fields in which emotion analysis is applied.

Further, emotion analysis on literature can be considered to be more challenging than in other domains. Literature contains more indirect communications of emotions, different stylistic means that need to be considered. It is not sufficient to analyze emotion expressions in isolation. They need to consider (personality) traits of characters and the embedding into situations. Therefore, it is, even more than in other areas, important to develop computational models of emotion that do not work in isolation but in the context of semantic representations of the story, the social network of characters, and many other components.

Finally, we believe that the integration of psychological models into computational approaches in literature studies is more important than in other domains. Literature contains representations of whole worlds; the depictions are more comprehensive than in news articles or social media. This also requires a deeper understanding of described social processes and (imagined) mental states.

And finally, the role of the experiencer of an emotion needs to be considered more than in other fields. While on Twitter analysis, we typically care about the emotion that the author of a message felt while writing it, we typically do not care about the emotion of the author of a novel, while writing it. Instead, we are faced with the more challenging task to attribute emotions to characters or even infer the emotions that might be developed by readers of a text.

In summary, we believe that the field of emotion analysis for literary studies is currently in its infancy. We have already seen sophisticated methods of role labeling, analyzing readers perception and characters experiences. However, the particular challenges in literary studies will in the future move from purely making use of methods that are (at least to some degree) developed outside of this domain, to a driving force that defines tasks to be solved.

## Acknowledgements

We thank Laura Ana Maria Oberländer, Sebastian Padó, and Enrica Troiano for fruitful discussions and the ZfDG team for their help in preparation of this article. This research has been conducted within the CRETA project which is funded by the German Ministry for Education and Research (BMBF) and partially funded by the German Research Council (DFG), projects SEAT (Structured Multi-Domain Emotion Analysis from Text, KL 2869/1-1). We further thank the anonymous reviewers for their helpful comments on an earlier version of this article.

| | | Features | | | | | | Models | | | | | | |
|---|---|---|---|---|---|---|---|---|---|---|---|---|---|---|
| | Citation | TF-IDF | Bag-of-words | Dictionary | Syntax | Embeddings | Annotation | SVM | Naïve Bayes | Decision Trees | Rules | PCA | Deep Learning | Other |
| Emotion classification | Yu (2008) | * | | | | | | * | * | | | | | |
| | Barros et al. (2013) | | | * | | | | | | * | | | | |
| | Reed (2018) | | | * | | | | | | | * | | | |
| | Zehe et al. (2016) | | | * | | | * | * | | | | | | |
| | Schmidt and Burghardt (2018) | | | * | | | * | | | | | | | |
| | Reagan et al. (2016) | | | * | | | | | | | | * | * | |
| | Samothrakis and Fasli (2015) | | * | * | | | | | | * | | | * | |
| | Kim et al. (2017a) | | * | * | | | | | | * | | | | |
| | Volkova et al. (2010) | | | * | | | * | | | | | | | |
| | Henny-Krahmer (2018) | | | * | | | | | | * | | | | |
| Temporal | Heuser et al. (2016) | | | | | | * | | | | | | | * |
| | Bruggmann and Fabrikant (2014) | | | * | | | | | | | * | | | |
| | Taboada et al. (2006, 2008) | | | * | | | * | | | | * | | | |
| | Chen et al. (2012) | | | * | | | | | | | * | | | |
| | Marchetti et al. (2014) | | | * | | | * | | | | * | | | |
| | Sprugnoli et al. (2016) | | | * | | | * | | | | * | | | |
| | Buechel et al. (2017) | | | * | | * | | | | | | | | |
| | Buechel et al. (2016) | | | * | | * | * | | | | | | | |
| | Leemans et al. (2017) | | | * | | | * | | | | * | | | |
| Network Analysis | Nalisnick and Baird (2013a,b) | | | * | | | | | | | * | | | |
| | Elsner (2012, 2015) | | | * | * | | | | | | * | | | |
| | Kim and Klinger (2018) | | | | | * | * | | | | | | * | * |
| | Barth et al. (2018) | | | * | | | | | | | * | | | |
| | Jhavar and Mirza (2018) | | | * | | | | | | | * | | | |
| | Egloff et al. (2018) | | | | | | * | | | | | | | * |
| | Rinaldi et al. (2013) | | | | | | | | | | | | | * |
| | Zhuravlev et al. (2014) | | | | | | | | | | | | | * |
| | Jarafi et al. (2016) | | | | | | | | | | | | | * |
| | Kim and Klinger (2019a) | | | | | | * | | | | | | | * |
| | Kim and Klinger (2019b) | | | | | * | * | | | | * | | * | |
| Other | Anderson and McMaster (1986) | | | * | | | | | | | * | | | |
| | ANderson and McMaster (1993) | | | * | | | | | | | * | | | |
| | Alm and Sproat (2005) | | | | | | * | | | | * | | | |
| | Mohammad (2011, 2012) | | | * | | | | | | | * | | | |
| | Klinger et al. (2016) | | | * | | | | | | | * | | | |
| | Kim et al. (2017b) | | | * | | | | | | | * | | | * |
| | Kakkonen and Galic Kakkonen (2011) | | | * | * | | | | | | * | | | |
| | Koolen (2018) | | | * | | | | | | | * | | | |
| | Kraicer and Piper (2019) | | | * | | | | | | | | | | * |
| | Morin and Acerbi (2017) | | | * | | | | | | | * | | | |
| | Bentley et al. (2014) | | | * | | | | | | | * | | | |

Table 1: Summary of characteristics of methods used in the papers reviewed in this survey.

## Bibliographic References

## List of Figures with Captions

Fig. 1: Plutchik’s wheel of emotions. In: Wikipedia, the free Encyclopedia: Robert

Plutchik. Article from 20.09.2019. [online] [Plutchik 2011. PD.]

Fig. 2: Circumplex model of affect: Horizontal axis represents the valence dimension; the vertical axis represents the arousal dimension. [Drawn after Posner et al. 2005.]